# Is novelty predictable?

Clara Fannjiang and Jennifer Listgarten
Department of Electrical Engineering and Computer Sciences, University of California, Berkeley
Email: {clarafy, jennl}@berkeley.edu

## Abstract

Machine learning-based design has gained traction in the sciences, most notably in the design of small molecules, materials, and proteins, with societal implications spanning drug development and manufacturing, plastic degradation, and carbon sequestration. When designing objects to achieve novel property values with machine learning, one faces a fundamental challenge: how to push past the frontier of current knowledge, distilled from the training data into the model, in a manner that rationally controls the risk of failure. If one trusts learned models too much in extrapolation, one is likely to design rubbish. In contrast, if one does not extrapolate, one cannot find novelty. Herein, we ponder how one might strike a useful balance between these two extremes. We focus in particular on designing proteins with novel property values, although much of our discussion addresses machine learning-based design more broadly.

## Table of Contents

# 1 Challenges in finding novelty with machine learning-based design

How can one find novelty, given only what is known? Focusing on machine learning-based protein design, herein we highlight challenges, current approaches to solving them, underexplored areas, and future directions.

The goal of protein design is to specify the sequence of a protein that satisfies a novel condition. There are three types of novel conditions a protein engineer may seek. One type of novelty is in the sequence—that is, one seeks a protein that differs in sequence from that of known proteins, but not necessarily in its structure or biochemical or biophysical properties. For example, one may seek a novel sequence for an enzyme in order to avoid a patent, or to serve as a different initial sequence for directed evolution. Much recent work in machine learning-based protein design focuses on sequence novelty, with applications ranging from gene therapy vectors (Bryant et al. 2021; Zhu et al. 2022) and antibodies (Shin et al. 2021) to signal peptides (Wu et al. 2020) and enzymes (Russ et al. 2020; Hawkins-Hooker et al. 2021; Repecka et al. 2021; Madani et al. 2023) A second type of novelty is in the structure—for example, when building a scaffold to contain and support a functional site (Correia et al. 2014; Sesterhenn et al. 2020; Wang et al. 2022).

The third type of novelty is in the biophysical or biochemical properties. That is, one seeks a protein with property values that have yet to be observed, which necessitates that the sequence, and likely the structure, are both also novel. Examples tackled with machine learning include enzymes with enhanced catalytic activity (Fox et al. 2007; Romero et al. 2013; Biswas et al. 2021; Greenhalgh et al. 2021; Fram et al. 2023), brighter fluorescent proteins (Brookes et al. 2019; Biswas et al. 2021; Stanton et al. 2022), optimized channelrhodopsins for optogenetics (Bedbrook et al. 2017, 2019) , and cell type-specific gene therapy vectors (Zhu et al. 2022). Herein, we focus on this third type of novelty. While machine-learning models can facilitate achieving this goal, the pursuit of novel property values is also the root cause of unique *in silico* challenges, the focus of our discussion.

The difficulty of these challenges depends on the extent of novelty sought. Is a protein with a melting point that is one degree higher, or that fluoresces slightly brighter at a similar wavelength, novel? Novelty is not a binary phenomenon. Rather, it exists on a spectrum, and the further one is on this spectrum the greater the challenges. At the extreme end is a notion of "radical novelty", which is not simply an improvement of an existing phenomenon, but rather a fundamentally different outcome. For example, one might refer to the design of a protein that catalyzes a reaction totally different from what known proteins can catalyze, as radical novelty. If the conditions one is interested in are not radically novel, then machine-learning based design may already be close to providing reliable solutions. Radical novelty is much more elusive.

We frame our discussion around three key tasks for machine learning-based design for novel property values. The first two are (i) learning a trustworthy model that makes predictions for the property of interest from, for example, a protein sequence and/or structure, and (ii) choosing what we call a design algorithm: an algorithm that consults the model to propose sequences intended to have the desired property values. Not all machine learning-based design approaches appear to explicitly comprise these two tasks, such as conditional generative modeling approaches. Nevertheless, these concepts and corresponding challenges are present under the hood. Although sometimes overlooked, there is also a third key task, that of (iii) uncertainty quantification. Quantifying uncertainty for the model's predictions helps a protein engineer understand what risk portfolio they are adopting when choosing a machine learning-based design strategy.

Our goal herein is not to comprehensively survey existing methods, nor to recast these into a unifying framework. Rather, we identify and discuss fundamental challenges, corresponding strategies, and underexplored areas that arise from the following dilemma: in order to find novel property values, any design algorithm must consider regions of sequence space away from the training data, but these regions are precisely where any learned model is least trustworthy.

## 1.1 Scope of protein design problems considered in this article

Sometimes the property of interest is sufficiently mediated by the protein's structure that the goal can be reframed as identifying a sequence that folds into a specific structure or substructure. A common example is when the goal is to bind to a target molecule where the structure of the desired complex is known with high resolution. Such structure-based design has traditionally been performed using biophysics-based modeling (Kuhlman and Bradley 2019), but is now increasingly performed with deep generative models (Norn et al. 2021; Dauparas et al. 2022; Hsu et al. 2022b; Watson et al. 2022); we refer the reader to (Ovchinnikov and Huang 2021; Pan and Kortemme 2021; Malbranke et al. 2023) for excellent reviews of these developments. However, there are many settings in which structure-based design is not a viable solution. In particular, how structural changes affect the property of interest is often not known with sufficient precision, with catalytic activity of enzymes being a classic example (Romero and Arnold 2009; Tokuriki and Tawfik 2009). Additionally, the property may be substantially mediated by conformational dynamics or quantum chemistry (Gao and Truhlar 2002; Faheem and Heyden 2014), neither of which can, at present, be readily captured by structure-based design. In these settings, one must rely additionally on directed evolution (Romero and Arnold 2009; Arnold 2018), and/or an approach based on a machine-learning model that leverages assay and/or evolutionary data relevant to the property of interest. We focus herein on this latter setting, which includes that of machine learning-assisted directed evolution (Wu et al. 2019; Wittmann et al. 2021).

Machine-learning models in this setting may be trained in a supervised fashion on sequences with labels from an experimental assay. Alternatively, some models may capture the property more implicitly, such as density models fit to families of protein sequences (Cheng et al. 2016; Figliuzzi et al. 2016; Hopf et al. 2017; Riesselman et al. 2018; Laine et al. 2019; Frazer et al. 2021; Shin et al. 2021; Trinquier et al. 2021) or to natural protein sequences more broadly (Alley et al. 2019; Madani et al. 2020; Meier et al. 2021; Ferruz et al. 2022; Notin et al. 2022a, b). The likelihood of these models, or approximations thereof, can provide useful information about protein properties. The conditional likelihoods of structure-conditioned generative models can similarly be used as predictions (Ingraham et al. 2022; Dauparas et al. 2022; Hsu et al. 2022b).

## 1.2 Overview of challenges

Almost surely, one does not have access to the true causal model for the property of interest, and without the true causal model, model predictions are almost surely wrong. They are particularly likely to be wrong given the shift between the distributions of training and designed sequences that are necessary when seeking novel property values. Despite the inability to access true causal models, predictive models can still be useful. How can one ensure that such models are useful for design? Motivated by this question, this article is organized into three sections, summarized next.

First, we ask: *how can one learn a trustworthy model for pursuing novel properties?* We discuss how different types of training data—such as evolutionary and assay-labeled—contain different types of noise and biases. We then discuss how the search for novel property values induces distribution shifts between the training and designed sequences that jeopardize how much one can trust the model. Finally, we discuss how to potentially mitigate such problems by infusing relevant biological knowledge into the inductive bias of the model, analogous to how the convolution operation in computer vision encodes fundamental knowledge about the translational invariance of objects in images.

Second, *how can one quantify uncertainty for design?* Because design necessitates making predictions for sequences far from the training data, the predictions can deviate wildly from the truth. It is therefore desirable to quantify the model's uncertainty, and thereby understand the risk one is incurring, before synthesizing and measuring designed sequences in the laboratory. To do so, one can choose from a variety of technical notions of uncertainty, each with its own strengths and limitations.

Finally, *what are the design algorithm considerations?* Given a fixed budget of sequences, how should the design algorithm place its bets so as to maximize the chances of finding a protein with the desired property values? One reasonable approach is for the design algorithm to take into account the model's uncertainty. Or, alternatively, perhaps one should quantify uncertainty jointly over the model and the design algorithm.

# 2 How can one learn a trustworthy model for pursuing novel conditions?

We start by describing how one's choices about the training data impact the model's predictions. We then discuss the distributional shifts that emerge from design, and finally, potential strategies to mitigate problems arising from these shifts.

## 2.1 Trade-off between quality and quantity of training data

To frame our discussion of how training data choices impact predictions, we appeal to the recently introduced bias-variance decomposition of prediction error described by Posani et al. (2022). While bias-variance decompositions have conventionally been used to help understand why different model classes yield different predictive performance (Geman et al. 1992; Hastie et al. 2001), Posani et al. (2022) repurpose the idea to understand how choices in curating homologous protein sequences affect how well the likelihood of a Potts model fit to those sequences provides a ranking of protein property values. Inspired by their work, we appeal to a similar framework to discuss the impact of training data more broadly on prediction error.

More concretely, consider the prediction error for a given test sequence, when averaged over predictions made from models trained on different random draws of data from a fixed distribution (Geman et al. 1992; Hastie et al. 2001).[1] This average prediction error can be decomposed into two parts. The first is a bias component, or the average difference between the prediction and true property value of the test sequence; intuitively, this component reflects how relevant the training data are for the prediction task at hand. The second, a variance component, captures how sensitive the prediction is to perturbations of the training data due to sampling. Intuitively, this component reflects the amount of information contained in the data set; more information is harder to perturb. Given equally relevant data, larger training data sets result in lower-variance predictions since the information contained in larger data sets is more robust to perturbations than that contained in smaller ones. These bias and variance terms exhibit a trade-off (Geman et al. 1992): if one chooses the training data in such a way as to improve one, then typically the other degrades. For example, increasing the amount of training data by incorporating data that is less relevant, such as homologous protein sequences that are more evolutionarily distant, results in increased bias, but decreased variance. Consequently, one can generally achieve different points on a trade-off curve by modulating the quantity versus quality of the training data. Next, we invoke this quantity-quality trade-off to gain insights into choosing both assay-labeled and evolutionary data.

**2.1.1 Assay-labeled data**

When choosing an experimental assay to label protein sequences, there is typically a trade-off between two extremes. On one end, there are high-quality, low-throughput experiments that yield measurements of a few dozen to a few thousand sequences at most, but which directly measure the biochemical or biophysical property of interest (Acker and Auld 2014; Markin et al. 2021). Training a model on the small amount of resulting data will result in low predictive bias but high predictive variance. On the other end, there are lower-quality, high-throughput assays that yield measurements of up to hundreds of thousands or even millions of sequences, but which only provide a biased proxy of the property of interest. A common form of the latter are sequencing-based assays known collectively as *deep mutational scanning*, in which a pool of sequences is subject to selection experiments that indirectly encourage sequences with the desired property values to become more abundant relative to other sequences (Fowler and Fields 2014; Wrenbeck et al. 2017). For example, to measure how well different enzyme sequences catalyze the biosynthesis of a compound necessary for a cell's survival, those enzymes can be expressed in a population of cells, whose subsequent survival rates reflect the catalytic efficiencies of the variants. From such data, one can derive a quantitative label for each sequence reflecting how much more abundant it became after the selection—for example, by computing ratios of counts of the sequence before and after the selection (Fowler et al. 2011; Rubin et al. 2017), although recent work shows how to improve such quantification using density ratio estimation (Busia and Listgarten 2023). Even though such labels indirectly inform us about the biophysical property of interest, they are generally biased since factors unrelated to the property can drive changes in sequence abundance (Song et al. 2021). Using them as training data consequently introduces predictive bias, albeit with low predictive variance due to the large data sets.

---

[1] Note that for the purposes of our discussion, we do not consider randomness in the test data, as done in some bias-variance analyses in the machine learning literature (Bishop 2007).

In either the high-quality-low-throughput or low-quality-high-throughput settings, it can be fruitful to augment assay-labeled data with evolutionary data—that is, increasing quantity at the expense of quality (Alley et al. 2019; Biswas et al. 2021; Hsu et al. 2022a). The blend of data types that achieves the optimal trade-off between quantity and quality will be different for each protein design effort. Ultimately, empirical assessments will dictate the choice, although these may be guided at times by theoretical results such as in Posani et al. (2022), discussed next.

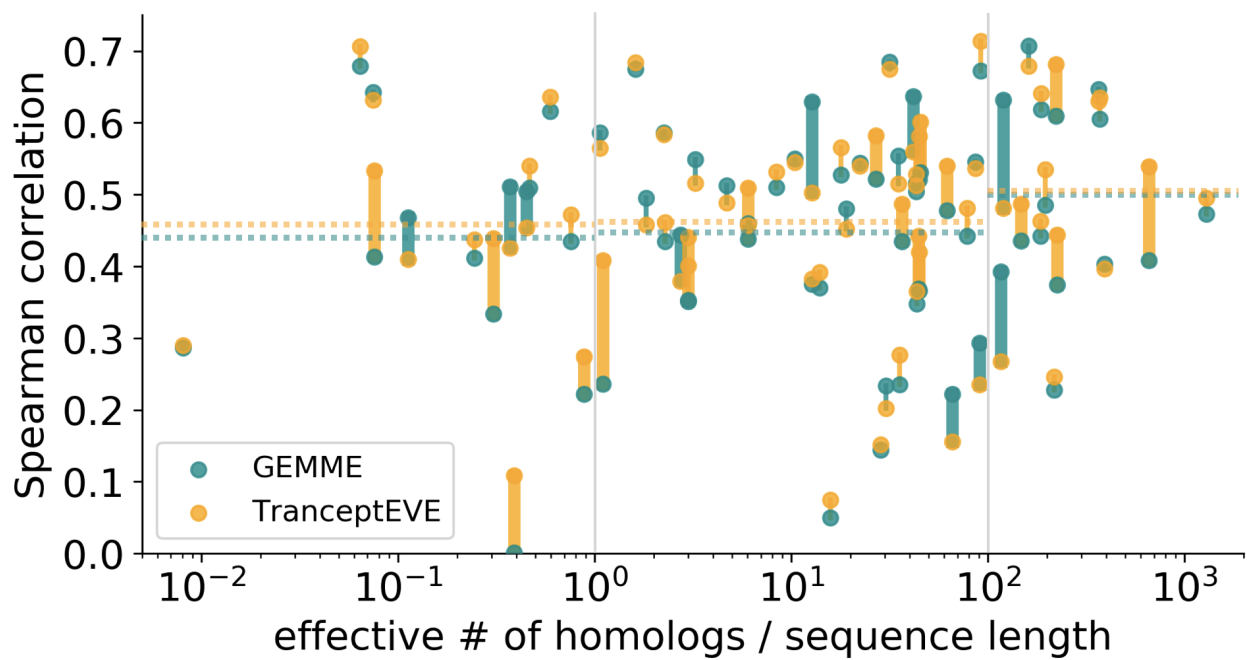

**Figure 1. Protein property predictive performance of the top two methods on the ProteinGym substitution benchmark.** All data in this plot are from Notin et al. (2022a, b) and the affiliated ProteinGym substitution benchmark at https://www.proteingym.org/substitutions. As of this writing, the two top-performing methods are one that combines pan-protein and family-specific models (TranceptEVE (Notin et al. 2022b), orange dots) and one family-specific method (GEMME (Laine et al. 2019), teal dots). The methods' performances on each ProteinGym data set (each teal and orange dot) is measured by Spearman correlations between the predictions and labels (higher is better). To improve visualization, the magnitude of the difference between the methods is shown by a solid vertical line. Its color denotes the winning method for that data set, and the line is thick if the difference is greater than 0.05 and thin otherwise. For different data sets comprising mutants of the same wild type protein, the average correlation over those data sets is shown as a single dot. The same thresholds from Notin et al. (2022a, b) were used to categorize the amount of homologous data as low, medium, or high (demarcated by two solid gray vertical lines), and the average correlation within each of these categories is shown for GEMME and TranceptEVE (dashed horizontal teal and orange lines).

### 2.1.2 Homologous sequence data
One can also understand the effects of curating homologous sequences with respect to the bias-variance trade-off. The goal is to identify a set of sequences known to appreciably exhibit the property of interest, and then fit a density model to those sequences. The likelihood of a sequence under such a model, if correctly specified, should then correlate with its real-valued property of interest. However, there are often only a handful of proteins, if any, that are laboratory-verified to exhibit the property of interest above some threshold. The likelihoods of a density model fit to such a small number of protein sequences would have low bias as property predictions, but extremely high variance. To reduce this variance at the expense of increased bias, one can increase the quantity of training data by using heuristics to identify proteins that are likely—but not known with certainty—to exhibit the property. One popular heuristic is to include homologous protein sequences, or *homologs*: those that are evolutionarily related to a natural protein that appreciably exhibits the property of interest. If the homologs underwent the same selective pressure that gave rise to that property, then they too should exhibit it. However, one never knows the extent to which they experienced that selective pressure, if at all. Moreover, finding homologs is itself a non-trivial task, as evolution cannot be observed over the relevant time scales. One must instead leverage heuristic search algorithms based on sequence similarity (Altschul et al. 1997; Johnson et al. 2010). Nevertheless, a rich line of work has shown that the likelihoods of various density models (Cheng et al. 2016; Figliuzzi et al. 2016; Hopf et al. 2017; Riesselman et al. 2018; Shin et al. 2021; Frazer et al. 2021; Trinquier et al. 2021) and predictions based on phylogenetic trees (Laine et al. 2019) fit to such putative homologs can be correlated with the property values of protein variants, despite the predictive bias introduced by data of uncertain relevance (Qin and Colwell 2018; Weinstein et al. 2022; Posani et al. 2022).

Algorithms for heuristically identifying homologs look through databases of natural protein sequences for similar sequences, and have various hyperparameters that can also be interpreted as navigating a quantity-quality trade-off. In particular, hyperparameter settings that return a larger set of sequences may include proteins that are not actually homologs, whereas settings that are too conservative may miss some (Pearson 2013). Posani et al. propose methods for selecting optimal subsets of homologs based on their sequence distances to a wild type. Selection based on auxiliary information, such as what types of species the homologs are from, can also be useful heuristics for reducing predictive bias (Jagota et al. 2022). Creatively addressing the sources of predictive bias and variance we have discussed may be fruitful for improving "zero-shot" protein property prediction in settings where assay-labeled data are not yet available.

### 2.1.3 Pan-protein data

The protein modeling community has borrowed ideas from the field of natural language processing to train unsupervised sequence models on what we call *pan-protein* data (*i.e.*, all known natural proteins, spanning all known protein families and beyond). The likelihood of these models, or approximations thereof, can also serve as zero-shot predictions of protein property values (Meier et al. 2021; Hesslow et al. 2022; Nijkamp et al. 2022; Notin et al. 2022a, b). In principle, learning a density over all these proteins could result in an implicit mixture of family-specific "modes", in which each mode captures the distribution of a single protein family, similarly to family-specific models such as Potts models. However, depending on the inductive bias of the pan-protein model architecture, such a model is likely to share information between families. The extent of this information-sharing can be viewed as implicit navigation of the bias-variance trade-off. Moreover, a single trained pan-protein model likely navigates this trade-off differently for different protein families: the fewer the known homologs for a family, for example, the more information the model may have borrowed from other families. What is the nature of this shared information, and what determines which families contribute to it? Such questions should be investigated so as to make sense of how, why, and when such models may be providing benefits, which in turn could be used to improve them or other approaches.

As of this writing, the best approach for zero-shot property prediction blends pan-protein and family-specific models (Notin et al. 2022b), which suggests that pan-protein models are not yet achieving an optimal trade-off by themselves. Indeed, the best family-specific method (Laine et al. 2019) performs similarly well, particularly when homologs are abundant for the family of interest (Figure 1) (Notin et al. 2022a, b). Far more conspicuous than the margin between these two methods' average performance is how dramatically their performance—as well as the magnitude of the difference between their performances—varies over different protein families (Figure 1), which raises open and underexplored questions. How much do the sources of predictive bias and variance we have described account for this variation? Given a protein family of interest, is it possible to anticipate beforehand how accurate the predictions from family-specific versus pan-protein methods will be?

We anticipate that in the future, it will be possible to forgo curation of homologs entirely and develop strategies for training pan-protein models to automatically find more effective points on the bias-variance trade-off. However, as family-specific models can currently perform similarly and are much easier to learn—or require minimal learning (Laine et al. 2019)—their use may continue well into the future.

## 2.2 Accounting for design-induced distribution shift

So far our discussion has been about the training data, without considering the distribution of test proteins that will emerge from design. We now examine this topic more closely.

Since we seek a protein with novel biophysical or biochemical property values, it follows that the designed proteins must come from a different distribution than the training proteins. This phenomenon is referred to as a *distribution shift*, which in many machine learning settings is passively observed rather than purposely induced as in the design setting. One type of distribution shift inherent to machine learning-based design for novel property values is *feedback covariate shift* (Fannjiang et al. 2022). As a refinement of the more common covariate shift (Shimodaira 2000), feedback covariate shift additionally encompasses settings where test inputs are chosen based on the training data, such that they are dependent on it, rather than simply drawn independently from a different distribution. Due to this distribution shift, the model's predictions must be trustworthy over regions of protein space "away from" the training proteins—in particular, regions characterized by the distribution of designed proteins, as well as those the model examines en route to finding that distribution.

If one knew these distributions in advance, then one could deploy strategies for learning a model to be accurate over the test rather than training distribution (Shimodaira 2000; Sugiyama et al. 2008; Bickel et al. 2009; Gretton et al. 2009). The chicken-and-egg dilemma, however, is that one does not know the distribution of designed proteins until after the model has been learned; this is the "feedback" described by feedback covariate shift. Nevertheless, it is prudent to anticipate and try to account for plausibly relevant distribution shifts. For example, many design algorithms move through protein space in an iterative fashion to search for promising proteins, where each move induces, either implicitly or explicitly (Brookes et al.

2020), an intermediary distribution of proteins currently under consideration. After each move, one could relearn the model in lockstep with the design algorithm such that it is more accurate over the intermediary distribution (Fannjiang and Listgarten 2020).

Learning strategies that account for distribution shift do have a cost: intuitively, all involve reweighting the training data to better mimic the test distribution. The more the training and test distributions differ, the greater the variance of these weights over the training data (Cortes et al. 2010), which means that the model learns from a smaller effective amount of data. Strategies for mapping the training and test proteins into feature spaces in which the distribution shift is more mild (Rhodes et al. 2020; Choi et al. 2022), or tempering the weights so as to only partially account for the shift (Grover et al. 2019) may provide solutions.

Beyond changing how the model is learned, uncertainty quantification strategies and design algorithms can also take design-induced distribution shifts into account, as we will discuss. In general, problems arising from such shifts disappear to the extent that the model captures the true causal mechanism. One way to move toward this causal model is to incorporate relevant knowledge into the inductive bias of the model, as discussed next.

## 2.3 Incorporating informative inductive biases

In practice, the true causal model is never available. As a practical mitigation strategy, one can imbue the inductive bias of the model with broadly applicable knowledge regarding how amino acid sequences give rise to biophysical or biochemical properties. For example, there is evidence that this relationship is dominated by single-site effects, and that higher-order or *epistatic* interaction effects are sparse and typically decay with increasing order (Sailer and Harms 2017; Otwinowski et al. 2018; Poelwijk et al. 2019; Yang et al. 2019; Ballal et al. 2020; Brookes et al. 2022; Ding 2022). Similarly, models that generate or make use of protein structures should respect physical constraints and rotational symmetries (Ingraham et al. 2019; Jing et al. 2021; Ingraham et al. 2022).

Particularly when using high-capacity deep learning models, incorporating such domain-specific, yet task-independent knowledge into the model can reduce its degrees of freedom without ruling out useful parts of parameter space. Doing so can consequently improve the efficiency with which the model distills relevant information from the training data. Intuitively, encoding these types of knowledge is tantamount to having additional high-quality data, enabling a reduction in both predictive bias and variance. Critically, since such knowledge is not based on the training data, a model that incorporates it can better retain its accuracy further away from the training data.

Coherently integrating appropriate knowledge into the model is a case-by-case challenge. Sometimes, as in the case of rotational symmetries, such knowledge is absolute and should be enforced as a hard constraint as in (Cohen and Welling 2016; Geiger and Smidt 2022), although computational issues can remain challenging. In contrast, the sparsity and decay of epistatic interactions with increasing order are phenomena that hold to different extents for different protein properties, so should only be softly enforced. In one approach, Aghazadeh et al. (2021) leverage ideas and algorithms from compressed sensing to encourage sparse epistatic interactions in neural networks, although they use a binary alphabet and do not discriminate between low- and high-order effects. The main technical challenge in incorporating knowledge about epistasis is how to tractably compute the combinatorially large number of higher-order interactions between protein sites (Erginbas et al. 2023), or devise clever ways to avoid doing so.

So far we have discussed incorporating knowledge that would be useful for almost any protein property prediction task. However, one can also leverage task-specific biophysical knowledge. While the accuracy of learned models degrades further away from the training data, biophysics-based models crafted in a relatively data-free manner should have roughly uniform accuracy over protein space, even if learned models outperform them near their training data. To capitalize on this intuition, Nisonoff et al. (2022) developed an easily instantiable and computation-efficient formalism for cohesively blending information from biophysics-based models with Bayesian neural network models. A key component of their approach is the uncertainty of the predictions, the topic of our next discussion.

## 3 How can one quantify uncertainty for design?

Thus far, we have discussed how the quality of the model's predictions, especially away from the training data, is affected by choices regarding the training data and inductive biases of the model. However, beyond improving predictions, one can also quantify uncertainty over them to assess risk in designing new proteins. The precise notion of uncertainty quantification that is most useful for protein design is an open and underexplored question. One must decide what entity needs uncertainty quantification, what notion of uncertainty is suitable, and finally how to go about quantifying it.

First, what is the entity whose uncertainty needs to be quantified? An obvious candidate is the property prediction for any individual test protein, for which we will discuss both Bayesian and frequentist approaches. In designing a library of proteins, however, one may alternatively want to quantify uncertainty for the average, median, or other quantiles of the property values contained in the library, so as to reason about its collective performance. This idea was first proposed and tackled by Wheelock et al. (2022) under a Bayesian lens, and will be fruitful to continue investigating.

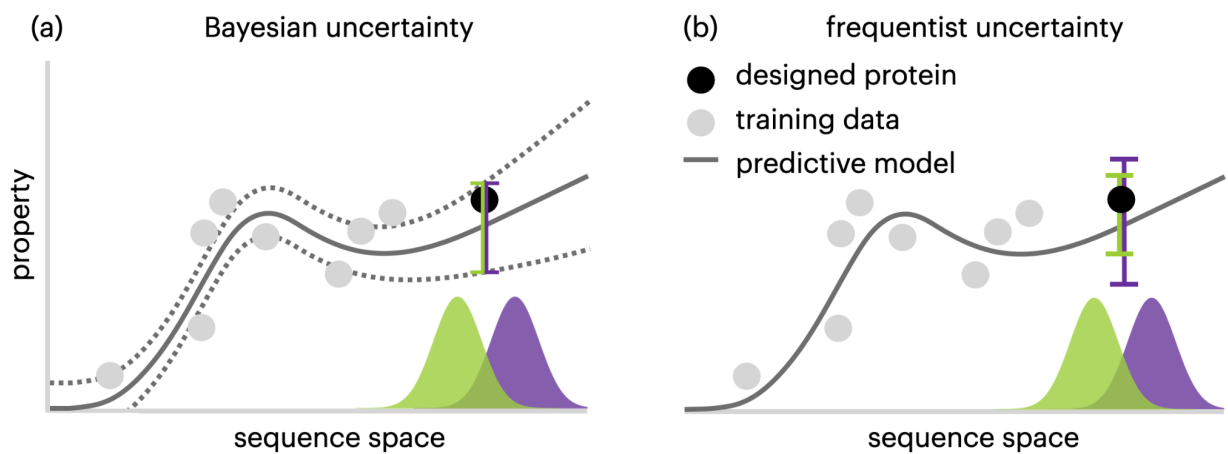

**Figure 2. Bayesian versus frequentist uncertainty quantification for design.** A predictive model is fit to training data (gray dots). A designed protein (black dot) is drawn from a design distribution; as examples of two different distributions, a purple one and a green one are shown. Confidence intervals for the designed protein are shown in corresponding colors. (a) Bayesian notions of uncertainty over a designed

protein tend to increase further from the training data, as illustrated by the distance between two standard deviations (dashed gray lines) above and below the mean (solid gray line) of the posterior predictive distribution. Given a designed protein, Bayesian notions of uncertainty do not change with the design distribution: for example, the interval bounded by the dashed gray lines is the same whether the protein was designed by sampling from the green distribution or the purple distribution. (b) In contrast, frequentist notions of uncertainty over a designed protein depend on the distribution it was drawn from. In particular, the confidence interval produced by a conformal prediction method that accommodates distribution shift will generally be smaller if the protein was sampled from a distribution closer to (green), rather than farther from (purple) the training distribution.

## 3.1 Bayesian uncertainty quantification

In the Bayesian framework, one specifies a prior distribution over the predictive model parameters that encodes the user's beliefs, as well as a likelihood model that provides, for each possible setting of the model parameters, the probability of the training data. The key Bayesian operation is to update the prior distribution to be more consistent with the evidence contained in the data, by reweighting each parameter setting according to the likelihood of the data under that setting. This update yields a posterior distribution which encodes updated beliefs about the predictive model parameters, which can in turn be used to derive a posterior distribution over the prediction for any test point, called the *posterior predictive distribution*. The variance or other measures of dispersion of the posterior predictive distribution are natural and commonly used notions of uncertainty for predictions.

An elegant aspect of the Bayesian framework is that different components can be interpreted as accounting for different sources of uncertainty (Kiureghian and Ditlevsen 2009). In particular, the likelihood model, if correctly specified, captures what is called *aleatoric uncertainty*, or uncertainty due to inherent stochasticity in the underlying causal mechanism being modeled. Aleatoric uncertainty never vanishes, even with infinite data. In contrast, the process of updating the prior based on the data captures reduction in *epistemic uncertainty*, or uncertainty due to lack of data, which vanishes with infinite data—at which point the prior has no influence over the posterior. One satisfying consequence is that posterior predictive distributions for test inputs further from the training inputs naturally have higher variance.

For exact, tractable computation of the posterior distribution, the Bayesian framework requires particular pairings of prior distributions and likelihood models, known as conjugate pairs, which may not be the ones that most accurately describe one's beliefs about the system. Nevertheless, these can work well in practice: Bayesian notions of uncertainty from Gaussian process regression models have been put to good use in protein engineering (Romero et al. 2013; Bedbrook et al. 2017, 2019; Greenhalgh et al. 2021; Rapp et al. 2023). Beyond the use of conjugate pairs, accessing the posterior requires sampling methods such as Markov Chain Monte Carlo methods (Neal 1996), or variational inference (Zhang et al. 2019; Gal 2016), which learns an approximation to the posterior that affords easy sampling. See Gal (2016) for a thorough study of tools for the latter in deep learning.

In contrast to frequentist approaches, Bayesian notions of uncertainty do not generally satisfy any guarantees about their relationship with the true value of the entity. For example, one might seek a guarantee that the true property value is less than the 90$^{th}$ percentile of the posterior predictive distribution 90% of the time—a notion referred to as *calibrated* uncertainty (Platt 1999). As the Bayesian framework centers around subjective beliefs (Gelman 2009; Fortuin 2022)—encoding and updating them—it does not yield such guarantees; in fact, such guarantees are antithetical to the Bayesian paradigm (Rubin 1984; Little 2006). In contrast, the frequentist framework, discussed next, explicitly seeks such guarantees.

## 3.2 Frequentist uncertainty quantification

The goal of frequentist uncertainty quantification is to produce uncertainty estimates that satisfy probabilistic notions of correctness. As an example, we will focus on *conformal prediction* methods, which yield confidence sets for test points that are guaranteed to contain the true labels with high probability, for any predictive model (Vovk et al. 2005; Lei et al. 2016; Angelopoulos and Bates 2023). Such *coverage* guarantees rely on assumptions about the relationship between the training and test data, such as being exchangeable (*e.g.*, independently and identically distributed), or related by way of particular distribution shifts.

The basic intuition with which to understand conformal prediction is that, if one has "calibration data"—held-out validation data from the same distribution as the test data—then the prediction error on a test point comes from the same distribution as the prediction errors on the calibration data. Consequently, one can obtain guarantees on the test error by assessing the calibration errors. However, we may not have access to such calibration data. Instead, for any test input, we can consider each possible value that its label could take on. For each such candidate label, we ask if the test input paired with that candidate label together "look like" they came from the same distribution as the training data, according to a frequentist hypothesis test. If they do, then this value is included in the confidence set for that test point. The confidence set that comprises all such values gives coverage: it contains the true test label with a probability determined by the hypothesis test's significance level.

The argument thus far holds when the training and test data are from the same distribution, or, more generally, exchangeable. Fannjiang et al. (2022) extend this framework to obtain coverage under feedback covariate shift, by devising the appropriate hypothesis test, building upon a recent body of work generalizing conformal prediction to various distribution shifts (Tibshirani et al. 2019; Cauchois et al. 2020; Gibbs and Candes 2021; Podkopaev and Ramdas 2021; Barber et al. 2022). Notably, the precise feedback covariate shift that emerges in design is dictated jointly by the design algorithm, predictive model, and training data. Consequently, the confidence set constructed for any given test protein depends on the design algorithm, not just the model and the training data. In contrast, a Bayesian predictive posterior for any given test protein depends on the model and training data but not on the design algorithm (Figure 2). Regardless, similar to the Bayesian setting, the more the training and design distributions differ, the larger the confidence sets produced by conformal prediction, capturing the intuition that predictions should be more uncertain.

Coverage guarantees are non-asymptotic, meaning they hold for any amount of training data. However, coverage is not a guarantee on the confidence set for any particular test protein—indeed, such *conditional coverage* guarantees are impossible without making strong assumptions about the sequence-property relationship (Vovk 2012; Foygel Barber et al. 2021). Rather, coverage describes what happens in expectation over random draws of both the training and test data from their respective distributions. Consequently, it may not be appropriate to use these methods to make choices about individual proteins. Rather, these methods are potentially most useful for selecting a design algorithm or its hyperparameters, as discussed in the next section on design algorithm considerations.

**3.2.1 Density ratio estimation: an important tool for handling distribution shift**
An additional limitation of conformal prediction methods in the design setting is their reliance on the density ratio between the training and the designed input distributions—that is, the ratio of the densities of these two distributions—in order to characterize the distribution shift (Tibshirani et al. 2019; Fannjiang et al. 2022). In fact, these ratios are also key to a panoply of learning strategies that use them to reweight training data in order to account for distribution shift (Shimodaira 2000; Sugiyama et al. 2008; Bickel et al. 2009; Gretton et al. 2009).

If both the training and design input distributions have known closed-form densities, then one can simply compute this ratio. For example, training sequences generated with library construction protocols such as error-prone polymerase chain reaction, degenerate codons, or recombination (Neylon 2004) can be framed as sampling from an explicit sequence distribution. Some design algorithms also prescribe sampling from a distribution with a closed-form density, such as a Potts model (Russ et al. 2020; Fram et al. 2023) or a library whose parameters have been set such that sequences with desirable predictions are more likely to be sampled (Weinstein et al. 2022; Yang et al. 2023; Zhu et al. 2022). However, in general this is not the case. For example, if training data comprise homologs, previously designed proteins, proteins gathered from different literature sources, or any combination thereof, then their distribution does not have a closed-form density. Similarly, most design algorithms only implicitly induce a design distribution.

As a naive solution, one could estimate the training and design distributions separately and take the ratio of their densities. However, the modeling choices that are best for estimating individual densities may not be the best for estimating ratios of different densities; in particular, this approach does not account for the nature of how the two distributions differ. Generally, estimation of density ratios is statistically fragile in that it becomes increasingly high-variance the more the distributions differ, particularly due to the high dimensionality of protein sequence space. The rich literature on density ratio estimation presents a wealth of alternative strategies that mitigate this problem (Sugiyama et al. 2012), which are ripe for further investigation in the context of machine learning-based design (Stanton et al. 2023). These strategies learn parametric forms of the density ratio by, for example, learning a classifier for distinguishing samples from the two distributions (Qin 1998; Bickel et al. 2009; Gutmann and Hyvärinen 2010), or minimizing various objectives quantifying how well the two distributions agree, after one of them is transformed by the density ratio estimate (Huang et al. 2006; Nguyen et al. 2007; Sugiyama et al. 2008; Gretton et al. 2009; Kanamori and Hido 2009; Sugiyama et al. 2012). These approaches explicitly or implicitly learn which features are most useful for modeling the differences between the two distributions, and which are irrelevant and can be ignored. Consequently, they can be more statistically efficient than the naive approach based on estimating the two densities separately.

Finally, a unique advantage of the design setting is that one can generate as many *in silico* designed sequences as desired for accurate density ratio estimation, provided the design algorithm is not too computationally costly.

## 3.3 Uncertainty quantification for models of evolutionary data

As a final note, an open and underexplored question is how to best quantify uncertainty when using the likelihoods of density models fit to evolutionary data as property predictions. Since likelihood lives in entirely different units from the property of interest and is therefore only correlated with it at best, it is unclear what entity one should quantify uncertainty for to facilitate design. Riesselman et al. (2018) invoked Bayesian notions of uncertainty over the model parameters to improve the correlation between likelihoods and property values. It might also be fruitful to quantify notions of uncertainty for this correlation, but it is not clear how a protein engineer could make use of such uncertainty to improve design.

# 4 What are the design algorithm considerations?

So far we have discussed considerations for learning a model whose predictions are as trustworthy as possible, and how to quantify its predictive uncertainty, particularly in the face of the distribution shifts induced by machine learning-based design. We now discuss considerations for choosing the design algorithm: the algorithm that leverages the model and associated uncertainty to propose protein sequences to measure in the laboratory. The goal may be to propose proteins either as the final set of designed proteins aspiring to achieve the desired property values, or to update the predictive model in an iterative manner.

As a concrete example, one simple design algorithm is as follows: pick some initial protein sequence; out of all possible single mutants of this sequence, choose the one with the most desirable predicted property value; repeat this step with the new mutant until a computational budget is exhausted, or the predictions suggest that the desired condition has been achieved. Alternatively, a design algorithm could entail sampling from a generative model—conditional or otherwise—that puts more weight on sequences with promising predicted property values (Brookes et al. 2019; Zhu et al. 2022; Weinstein et al. 2022), a density model fit to homologs, such as a Potts model or variational autoencoder (Russ et al. 2020; Hawkins-Hooker et al. 2021; Fram et al. 2023), or a pan-protein model (Ferruz et al. 2022; Madani et al. 2023). These examples show that the design algorithm can be either intertwined with or decoupled from the model. Either way, the design algorithm dictates how the design distribution will shift relative to the distribution of the training data. As such, how should one select the design algorithm so as to maximize the chance of success?

Because design algorithms employed to seek novel property values must consider sequence space away from the training data, an inherent trade-off arises between "exploration"—proposing sequences whose predictions appear to meet the desired condition but have high uncertainty, versus "exploitation"—proposing sequences with more confident but less promising predictions. Fundamentally, novelty-seeking is about navigating this dilemma as effectively as possible.

## 4.1 Bayesian optimization

Bayesian optimization is one richly studied framework for tackling this dilemma in the context of iterative rounds of design, so-called because the predictive model is updated in a Bayesian manner after every round of data collection (Snoek et al. 2012; Shahriari et al. 2016). Instantiated in protein engineering terms, Bayesian optimization requires that one specify an acquisition function, a function over protein space whose maximum dictates which protein should be measured next in the laboratory. Acquisition functions typically incorporate both a protein's predicted property and the associated uncertainty. For example, the commonly used *upper confidence bound* algorithm can be understood as Bayesian optimization with an acquisition function comprising a weighted sum of the posterior predictive mean and standard deviation, where the weight is specified by the user and controls the exploration-exploitation trade-off. After using an optimization algorithm to maximize the acquisition function and identify which protein to measure next in the laboratory, the new data can be used to update the predictive model, and the process is iterated. In batch Bayesian optimization, one can design protein libraries using an acquisition function that is over a set of protein sequences (Azimi et al. 2010; Shah and Ghahramani 2015; Wu and Frazier 2016; Daxberger and Low 2017; Gonzalez et al. 2016; Yang et al. 2019).

Bayesian optimization approaches, especially the upper confidence bound algorithm instantiated with Gaussian process regression models (GP-UCB), have been successfully used in a variety of protein engineering campaigns (Romero et al. 2013; Bedbrook et al. 2017, 2019; Greenhalgh et al. 2021; Rapp et al. 2023). Justifications for the use of GP-UCB and its batched variants sometimes invoke their theoretical properties, such as efficient rates of convergence in finding the global optimum for many functions (Srinivas et al. 2010; Desautels et al. 2014). However, it is unclear how practically informative such rates are, particularly since protein engineers typically want to design proteins in as few rounds as possible, aspirationally in just a single round. Consequently, further analysis and development of algorithms tailored specifically for the low- or single-round setting could be fruitful (Chan et al. 2021).

The form of many common acquisition functions assumes that the predictive model yields a conditional density of the label given the input, such as of the binding affinity to a target molecule given the protein sequence. However, when there is no assay-labeled data to begin with, one often uses the likelihood of a density model fit to homologous or pan-protein data to instead rank proteins. In such cases, it is not straightforward how to instantiate common acquisition functions. Moreover, maximizing the acquisition function—which is not in general concave or otherwise friendly to optimization—is itself nontrivial (Wilson et al. 2018). Finally, if one wants frequentist-style guarantees on the output of the design algorithm, then Bayesian optimization may not be the most amenable paradigm (though see Stanton et al. (2023), who integrate the two by modulating the Bayesian predictive posterior with conformal confidence sets). For any of these reasons, alternative design algorithms, discussed next, can be employed.

## 4.2 Beyond Bayesian optimization

A number of design algorithms that do not subscribe to the Bayesian optimization paradigm have been developed and used successfully. Many of these can be framed as finding protein sequences that optimize some sort of acquisition function; however, since they do not involve any update of the predictive model, Bayesian or otherwise, they are neither performing Bayesian optimization nor intended for multi-round design. The goal of the "acquisition function" in these approaches is to navigate the same exploration-exploitation tradeoff already discussed, albeit in the setting where collected data will be final and cannot be used to inform another round of proposed proteins.

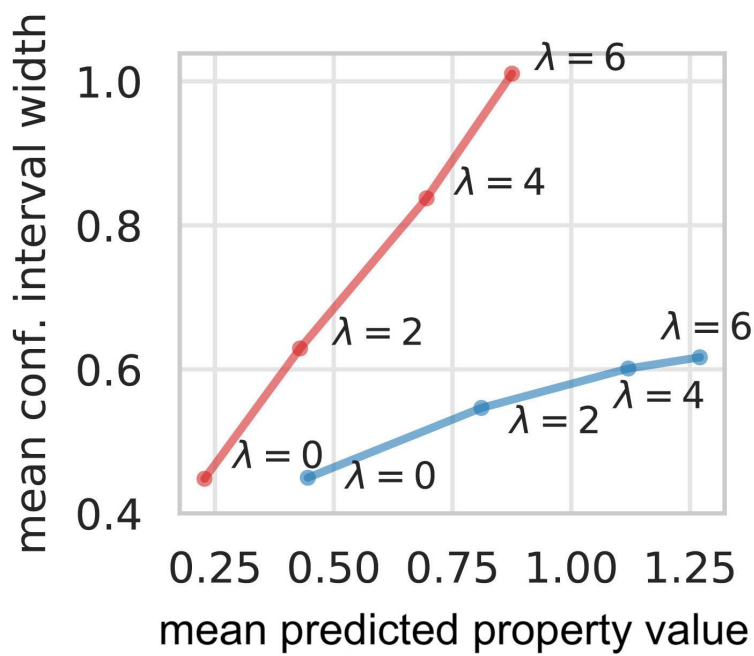

**Figure 3. Design algorithm hyperparameter selection with conformal prediction.** The "inverse temperature" hyperparameter, $\lambda$, controls the entropy of the design distribution (higher values lead to lower entropy). Higher values also correspond to higher average predicted property values (*x*-axis) and greater predictive uncertainty on the designed proteins, as measured by the width of confidence intervals produced by conformal prediction (*y*-axis, higher is more uncertain). The trade-off is shown for two different protein design goals: achieving brighter red and blue fluorescence (red and blue lines). One can access much higher predicted brightness for blue fluorescence without incurring as much predictive uncertainty, compared to red fluorescence (*i.e.*, the blue line is less steep than the red line). One explanation for this behavior is that the replicate red fluorescence measurements from Poelwijk et al. (2019) are noisier and hence inherently harder to predict. Credit: Fannjiang et al. (2022)

Such design algorithms include those that sample designed proteins using an Estimation of Distribution Algorithm (Brookes et al. 2019, 2020; Fannjiang and Listgarten 2020), a black-box optimization strategy that can be used to find the parameters of a sequence distribution with, say, maximal expected property value. Others use gradient-based methods with differentiable predictive models to design sequences with desirable predicted values (Killoran et al. 2017; Bogard et al. 2019), or start with an initial set of proteins and iteratively introduce and accept mutations based on their predicted property values (Fox et al. 2007; Sinai et al. 2020; Bryant et al. 2021). When mutations are chosen in a suitable manner, the latter approach is equivalent to Markov chain Monte Carlo sampling from an explicit distribution (Biswas et al. 2021). Although these methods cannot be analyzed through the theoretical lens of Bayesian optimization, one can wrap frequentist formalisms for uncertainty quantification around them to get guarantees on their outputs. For example, as discussed in the previous section, Fannjiang et al. (2022) generalize conformal prediction to the setting of machine learning-based design. By framing any design algorithm as a mapping from training data to a design distribution—even if only implicitly—this method can provide frequentist uncertainty guarantees for the output of any combination of design algorithm, predictive model, and (independently and identically distributed) training data. This flexibility means that one can incorporate whatever heuristics, intuitions, or constraints one desires, such as various mechanisms for encouraging designed sequences to remain in regions where the predictive model is trusted (Brookes et al. 2019; Linder et al. 2020; Biswas et al. 2021; Fram et al. 2023), and still obtain the same type of guarantees.

Although not the original motivation for conformal prediction, these methods can be used as a tool for design algorithm selection, including how to set hyperparameters of a design algorithm in an informed manner. For example, many design algorithms have a hyperparameter that can be thought of as navigating the exploration-exploitation tradeoff. Common examples include a "temperature" hyperparameter that controls the entropy of the design distribution (Russ et al. 2020; Biswas et al. 2021; Zhu et al. 2022), and a prediction threshold hyperparameter in genetic algorithms (Sinai et al. 2020). In general, it is unclear how to set these and other hyperparameters. However, the tool of conformal prediction can help one gauge how different values of the hyperparameter trade off between desirable but conflicting goals—for example, the trade-off between how high the average predicted value is, and frequentist uncertainty about the predictions. Plotting such trade-offs can guide protein engineers in selecting a hyperparameter value that they believe achieves an acceptable compromise, or multiple such values to achieve a risk portfolio (Figure 3).

As one moves along the trade-off in Figure 3, not only do the predicted property values and their uncertainties change, but also the diversity of the designed sequences: the higher the value of the hyperparameter, the lower the entropy of the design distribution. We next discuss the topic of sequence diversity in protein design.

## 4.3 Should sequence diversity be an explicit consideration?

In any setting where a batch of sequences, rather than just a single sequence, is designed at once, a common concern might be how to propose a sufficiently diverse set of sequences so as to maximize the chance of achieving the desired condition. It turns out that this explicit goal of sequence diversity is often beside the point. If the real goal is to achieve novel property values,[2] then, provided there is a suitable notion of uncertainty for the property predictions, choosing sequences in a way that accounts for that uncertainty is the principled strategy for exploring sequence space. For example, in the batch Bayesian optimization setting, diversity is effectively baked into the acquisition function. Once the acquisition function has been decided, it implicitly determines the appropriate notion of diversity in the resulting proposed batch of sequences. Analogously, if one uses conformal prediction to select hyperparameter settings to navigate the exploration-exploitation trade-off (Figure 3), then the appropriate notion of sequence diversity is dictated by the selected hyperparameter value. From these two examples, we see that in a design approach that sensibly accounts for the user's notion of uncertainty about the property predictions, the correspondingly appropriate notion of sequence diversity emerges naturally.

As just described, ideally the design algorithm accounts for predictive uncertainty such that it implicitly yields an effective notion of sequence diversity. However, in practice, there may not be a clear notion of predictive uncertainty that can be leveraged—for example, when using the likelihoods of a density model fit to evolutionary data to rank protein sequences. It is in such cases, or when one is unsure of what uncertainty quantification approach to use, but finds notions of sequence diversity easier to specify, that it may be effective to introduce property-agnostic sequence diversity metrics into the design algorithm. Examples include the entropy of the designed sequence distribution, or Hamming or BLOSUM-based distances between protein sequences (Angermueller et al. 2020; Linder et al. 2020). One could also consider generalizing such distances to account for higher-order interactions between sequence positions, or otherwise incorporate knowledge about the sequence-property relationship into the metric. As a side note, one should not use entropy to assess the diversity of a finite set of sequences, as the entropy of any set of unique sequences is the same regardless of how different they are from each other.

---

[2] In contrast, as noted in the introduction, if the goal is to find proteins with novel sequences—but not necessarily novel properties—then sequence diversity should be explicitly considered.

# 5 Prediction and uncertainty about the future of machine learning-based protein design

With the recent rise of large generative models for "designing" written compositions and images with astonishing quality, protein engineers have been motivated to follow a similar path by training generative models on large and diverse sets of protein sequences and/or structures. Will such efforts mean that the challenges discussed herein will soon become irrelevant? Important distinctions exist between the setting of protein design and that of language and image generation, although the practical implications of such distinctions remain to be seen. Still, let us consider them.

In large language models and their corresponding dialogue systems (OpenAI 2023), the training data, queries, and outputs all live in the same space: they are all instantiated in terms of language tokens, or, roughly, words. Correspondingly, the information that the model needs can be extracted entirely from textual data. Now consider a similar type of model, trained only on protein sequences and no other information. Could such a model know which proteins fluoresce at a particular wavelength with a particular brightness? No, that information lies in a space unfamiliar to the model—that of real-valued wavelength and brightness, not sequence. The fact that the number of known natural protein sequences is steadily rising due to plummeting sequencing costs does not reveal this information. In order for a model to know about the biochemical and biophysical properties of any given sequence, information from this space, such as measurements from laboratory experiments—a different modality altogether—must also be provided to the model during training. Databases of natural protein sequences do contain (mostly qualitative) annotations about biophysical and biochemical properties, but the quality of such annotations varies greatly. Although some are from published experiments, many are based on heuristics such as propagating annotations from similar sequences, which can be error-prone in a way not attenuated by scale (Brenner 1999; Schnoes et al. 2009; Radivojac et al. 2013). Consequently, fully automated protein design is currently bottlenecked by a lack of informative data from the relevant space. These data could come from laboratory experiments, or sufficiently accurate predictive models or simulations thereof. We anticipate that this bottleneck will not be overcome in the near future.

In recent work, Madani et al. (2023) incorporated annotations from a variety of sources into training a pan-protein generative model, and then conditioned on protein family annotations based on sequence homology (Finn et al. 2016) to generate functional proteins from specific families. However, they also fine-tuned the model on abundant homologous sequences from those families. Notably, when such homologous sequences are available, functional proteins have been successfully designed without access to pan-protein data (Russ et al. 2020;

Hawkins-Hooker et al. 2021; Repecka et al. 2021; Fram et al. 2023). Although Madani et al. (2023) find that one particular family-specific method (Figliuzzi et al. 2018; Russ et al. 2020) did not yield functional proteins for their families of interest, we discussed earlier how the relative performance of pan-protein and family-specific methods varies dramatically across protein families (Figure 1). Until we better understand the factors that drive this variation, general advantages of the use of pan-protein data and models remain unclear.

Even if there should eventually be enough suitable annotated sequence data to mimic the successes of large, multi-modal models on images and text, there remains a question of how well such models could achieve novel property values, or "radically novel" properties altogether. Claims of dialogue agents exhibiting "emergent behaviors" suggest that these models may be capable of finding novelty, although such claims have also been questioned (Schaeffer et al. 2023) due to the fact that emergent behavior has not been well-defined at a technical level. Consequently, it is difficult to reason about how relevant these supposed emergent behaviors are to the world of protein design. Let us see what emerges—we cannot predict the future!

## Author contributions
Whatever you agree with, CF wrote. Whatever you disagree with, JL wrote.

## Acknowledgments
We are grateful to Akosua Busia, Albert Fannjiang, Hanlun Jiang, Hunter Nisonoff, and Micah Olivas for providing compelling feedback, literature pointers, and snacks.

https://www.biorxiv.org/content/10.1101/2020.03.07.982272v2 (Accessed May 16, 2023).

Malbranke C, Bikard D, Cocco S, Monasson R, Tubiana J. 2023. Machine learning for evolutionary-based and physics-inspired protein design: Current and future synergies. *Curr Opin Struct Biol* **80**: 102571.

Markin CJ, Mokhtari DA, Sunden F, Appel MJ, Akiva E, Longwell SA, Sabatti C, Herschlag D, Fordyce PM. 2021. Revealing enzyme functional architecture via high-throughput microfluidic enzyme kinetics. *Science* **373**. https://science.sciencemag.org/content/373/6553/eabf8761/tab-figures-data (Accessed July 27, 2021).

Meier J, Rao R, Verkuil R, Liu J, Sercu T, Rives A. 2021. Language models enable zero-shot prediction of the effects of mutations on protein function. *Adv Neural Inf Process Syst* **34**: 29287–29303.

Neal RM. 1996. *Bayesian Learning for Neural Networks*. Springer New York.

Neylon C. 2004. Chemical and biochemical strategies for the randomization of protein encoding DNA sequences: library construction methods for directed evolution. *Nucleic Acids Res* **32**: 1448–1459.

Nguyen X, Wainwright MJ, Jordan M. 2007. Estimating divergence functionals and the likelihood ratio by penalized convex risk minimization. *Adv Neural Inf Process Syst* **20**. https://proceedings.neurips.cc/paper/2007/hash/72da7fd6d1302c0a159f6436d01e9eb0-Abstract.html.

Nijkamp E, Ruffolo J, Weinstein EN, Naik N, Madani A. 2022. ProGen2: Exploring the Boundaries of Protein Language Models. *arXiv [csLG]*. http://arxiv.org/abs/2206.13517.

Nisonoff H, Wang Y, Listgarten J. 2022. Augmenting Neural Networks with Priors on Function Values. *arXiv [csLG]*. http://arxiv.org/abs/2202.04798.

Norn C, Wicky BIM, Juergens D, Liu S, Kim D, Tischer D, Koepnick B, Anishchenko I, Foldit Players, Baker D, et al. 2021. Protein sequence design by conformational landscape optimization. *Proc Natl Acad Sci U S A* **118**. http://dx.doi.org/10.1073/pnas.2017228118.

Notin P, Dias M, Frazer J, Hurtado JM, Gomez AN, Marks D, Gal Y. 17--23 Jul 2022. Tranception: Protein Fitness Prediction with Autoregressive Transformers and Inference-time Retrieval. In *Proceedings of the 39th International Conference on Machine Learning* (eds. K. Chaudhuri, S. Jegelka, L. Song, C. Szepesvari, G. Niu, and S. Sabato), Vol. 162 of *Proceedings of Machine Learning Research*, pp. 16990–17017, PMLR.

Notin P, Van Niekerk L, Kollasch AW, Ritter D, Gal Y, Marks DS. 2022. TranceptEVE: Combining Family-specific and Family-agnostic Models of Protein Sequences for Improved Fitness Prediction. *bioRxiv* 2022.12.07.519495. https://www.biorxiv.org/content/10.1101/2022.12.07.519495v1?rss=1 (Accessed May 24, 2023).

OpenAI. 2023. GPT-4 Technical Report. *arXiv [csCL]*. http://arxiv.org/abs/2303.08774.

Otwinowski J, McCandlish DM, Plotkin JB. 2018. Inferring the shape of global epistasis. *Proc Natl Acad Sci U S A* **115**: E7550–E7558.
25

EF, Nowakowski TJ, et al. 2022. Optimal trade-off control in machine learning-based library design, with application to adeno-associated virus (AAV) for gene therapy. *bioRxiv* 2021.11.02.467003. https://www.biorxiv.org/content/10.1101/2021.11.02.467003v3 (Accessed May 5, 2023).